\documentclass[conference]{IEEEtran}
\IEEEoverridecommandlockouts
\usepackage{cite}
\usepackage{amsmath,amssymb,amsfonts}
\usepackage{algorithmic}
\usepackage{subcaption}
\usepackage{graphicx}
\usepackage{textcomp}
\usepackage{xcolor}
\usepackage{booktabs}
\usepackage{tikz}
\usetikzlibrary{tikzmark}
\def\BibTeX{{\rm B\kern-.05em{\sc i\kern-.025em b}\kern-.08em
    T\kern-.1667em\lower.7ex\hbox{E}\kern-.125emX}}
\begin{document}

\title{Enhancing Image Classification with Augmentation: Data Augmentation Techniques for Improved Image Classification\\

}

\author{
  \IEEEauthorblockN{Saroj Kumar*}
  \IEEEauthorblockA{
    \textit{Schulich School of Engineering} \\
    \textit{University of Calgary}\\
    Calgary, Canada \\
    saroj.kumar@ucalgary.ca
  }
  \and
  \IEEEauthorblockN{Ugochukwu Esiowu}
  \IEEEauthorblockA{
    \textit{Schulich School of Engineering} \\
    \textit{University of Calgary}\\
    Calgary, Canada \\
    franklyn.esiowu@ucalgary.ca
  }
  \and
  \IEEEauthorblockN{Prince Asiamah}
  \IEEEauthorblockA{
    \textit{Schulich School of Engineering} \\
    \textit{University of Calgary}\\
    Calgary, Canada \\
    prince.asiamah@ucalgary.ca
  }
  \and
  \IEEEauthorblockN{
    \textit{}
    \textit{}
  }
  \and
  \IEEEauthorblockN{\hspace{8em} Oluwatoyin Jolaoso}
  \IEEEauthorblockA{
    \textit{\hspace{9em}Schulich School of Engineering} \\
    \textit{\hspace{9em}University of Calgary}\\
    \hspace{9em}Calgary, Canada \\
    \hspace{9em}toyin.jolaoso@ucalgary.ca
  }
}

\maketitle

\begin{abstract}
Convolutional Neural Networks (CNNs) serve as the workhorse of deep learning, finding applications in various fields that rely on images. Given sufficient data, they exhibit the capacity to learn a wide range of concepts across diverse settings. However, a notable limitation of CNNs is their susceptibility to overfitting when trained on small datasets. The augmentation of such datasets can significantly enhance CNN performance by introducing additional data points for learning. In this study, we explore the effectiveness of 11 different sets of data augmentation techniques, which include three novel sets proposed in this work. The first set of data augmentation employs pairwise channel transfer, transferring Red, Green, Blue, Hue, and Saturation values from randomly selected images in the database to all images in the dataset. The second set introduces a novel occlusion approach, where objects in the images are occluded by randomly selected objects from the dataset. The third set involves a novel masking approach, using vertical, horizontal, circular, and checkered masks to occlude portions of the images. In addition to these novel techniques, we investigate other existing augmentation methods, including rotation, horizontal and vertical flips, resizing, translation, blur, color jitter, and random erasing, and their effects on accuracy and overfitting. We fine-tune a base EfficientNet-B0 model for each augmentation method and conduct a comparative analysis to showcase their efficacy. For the evaluation and comparison of these augmentation techniques, we utilize the Caltech-101 dataset. The ensemble of image augmentation techniques proposed emerges as the most effective on the Caltech-101 dataset. The results demonstrate that diverse data augmentation techniques present a viable means of enhancing datasets for improved image classification.
\end{abstract}

\begin{IEEEkeywords}
Convolutional Neural Networks (CNNs), Deep Learning, Image Classification, Data Augmentation, Overfitting, EfficientNet-B0, Pairwise Channel Transfer, Novel Occlusion Approach, Novel masking approach, Caltech-101 Dataset \cite{b18}. 
\end{IEEEkeywords}

\section{Introduction}
Making machines learn requires data from which they can derive real-world representations. With recent advances in machine learning, which involve very large neural networks, training them has become a highly data-intensive task. This area of machine learning is known as Deep Learning, and it requires millions of data points. According to recent trends, the successful application of Deep Learning increasingly demands more data \cite{b17}. There are several methods to increase the size of a high-quality dataset; however, One of the simplest ways to enhance the quality of the dataset is through data augmentation.

\subsection{Data augmentation}
Data augmentation is the technique to artificially increase the diversity in a training dataset by using various types of transformations. Data augmentation techniques are used to improve the model’s ability to generalize well to new and unseen data and thus improve its robustness. By using data augmentation techniques, the existing dataset becomes large and more diverse. This helps in avoiding the overfitting of the model and learning features that are invariant to such transformations. For visual datasets such as RGB images, there are several common data augmentation techniques such as rotation, flip, translation, color jitter, blur, random erasing, etc \ref{b17}. As the name suggests, Rotation involves rotating the image by a certain angle which introduces rotation invariance in the model. Flip involves flipping the image horizontally or vertically which leads to orientation invariance in the model. Translation involves changing the position of the image horizontally or vertically which introduces location invariance in the model. In a similar fashion, color jitter, which involves changing the color of the image, causes lighting robustness in the model. Therefore, the model would be able to detect an object in an image even in difficult lighting conditions. Random erasing involves removing random patches of the image which introduces occlusion invariance in the model. All these image augmentation techniques lead to better performance of the model in terms of different metrics; however, if we can better learn the representation from the data with novel data augmentation techniques, it would lead to a significant improvement in different performance metrics.
\subsection{Proposed Augmentation}
In this work we propose three novel sets of data augmentation techniques. First set of data augmentation is called Pairwise chanel transfer. Second set of augmentation technique are novel occlusion approach and the third set of augmentation technique are novel masking approach
\subsubsection{Pairwise chanel transfer}
The first set of data augmentation techniques involves pairwise channel transfer in images, where a channel from a random image in the dataset is transferred into other images in the dataset. This process facilitates the cross-pollination of information between images and introduces adulteration, which promotes irrelevant content invariance in the images.
\subsubsection{Novel occlusion Augmentation}
The second set of image augmentation techniques introduces a novel occlusion approach, where objects within an image are occluded by random images from the dataset. This augmentation is expected to result in a model that is better at handling occlusions.
\subsubsection{Novel masking augmentation}
The final and third set of image augmentation techniques involves a novel masking approach, where the images and their contents are masked by horizontal, vertical, checkered, and circular stripes. Such masking is expected to lead to better representation and help prevent overfitting of the model.

In addition to these proposed augmentation techniques, we also investigate existing image augmentation methods, namely rotation, horizontal and vertical flips, resizing, translation, blur, color jitter, and random erasing. These techniques were tested on the Caltech-101 dataset \cite{b18}.

\subsection{Dataset Used}
This dataset comprises 101 different categories of images, with each category containing between 40 and 800 images. Based on the proposed and existing augmentation techniques, we have developed three different variants of the dataset. The first variant is a vanilla dataset with no data augmentation, totaling 9,146 images. The second variant incorporates the original dataset enhanced by existing data augmentation techniques, resulting in a total of 54,864 images. The third variant includes the original dataset combined with newly proposed data augmentation techniques, leading to a total of 73,153 images.

\subsection{Deep learning model used}
We train the base variant of the EfficientNet\_B0 deep neural network \cite{b16} on these three different variants of the dataset to assess the performance of the proposed data augmentation techniques. Introduced by Tan et al., EfficientNet\_B0 is a deep Convolutional Neural Network that is part of a study investigating the effects of width, depth, and resolution on the accuracy of neural networks. EfficientNet\_B0, the smallest of the proposed models, has the least width, depth, and resolution, which results in the fewest number of parameters. Among all the proposed deep learning models, this variant has 4,138,210 trainable parameters.

\section{Background}
Image augmentation and associated techniques have been foundational in the field of computer vision and machine learning. LeCun et al. \cite{b11} employed planar affine transformations, such as horizontal and vertical translations and scaling, to augment images for handwritten zip code recognition. This significantly improved test error rates in their seminal work, which introduced the Convolutional Neural Network (CNN) known as LeNet-5. Similarly, groundbreaking research known as AlexNet by Kirchofsky et al. \cite{b12} utilized techniques such as horizontal reflection and color jitter based on eigenvalues and eigenvectors for each channel, leading to a considerable increase in dataset size. These examples underscore the critical importance of image augmentation in training machine learning and deep learning models.

Recent advances in computer vision also reflect similar trends, particularly those involving transformer architecture \cite{b19} for image classification tasks. Here, several augmentation techniques, including random horizontal flipping and cropping of images, are employed to enhance the dataset, as detailed in \cite{b20}. Moreover, very recent work on image classification \cite{b21}, which achieves state-of-the-art classification accuracy on ImageNet \cite{b22}, employs data augmentation techniques like random horizontal flips and cropping. These tasks are essential as semantic labels are invariant to such transformations.

Data augmentation techniques are ubiquitous in computer vision, consistently leading to improved results. The significant impact of image augmentation is well-documented, prompting several researchers to propose novel techniques. For example, Cubuk et al. \cite{b23} introduced a method called AutoAugment. This technique selects an augmentation from a predefined set based on a policy that optimizes accuracy for a specific dataset. This policy also dictates the probabilities and magnitudes with which these augmentations are applied. When applied to the CIFAR-10 \cite{b24} dataset, AutoAugment achieved state-of-the-art results. Similarly, Dai et al. \cite{b25} proposed a technique that involves blending two images and then randomly erasing sections of the resultant image. This method improved performance by 1-2\% in image classification tasks.

In another innovative approach, Chen et al. \cite{b26} introduced GridMask, which involves masking the image with a grid of black patches at various scales. Their experiments showed that GridMask performed better on the ImageNet \cite{b22} dataset than the previously mentioned AutoAugment scheme.

DeVries et al. \cite{b27} proposed an intuitive data augmentation technique called cutout, where they randomly masked a square area of images. Upon experimentation, they found that this augmentation led to increased accuracy and robustness of the model. They also found that this data augmentation technique can be used in conjunction with other regularizers and augmentation techniques. When they applied this data augmentation technique to CIFAR-10 \cite{b24}, CIFAR-100 \cite{b28}, and SVHN datasets \cite{b29} by utilizing the current state-of-the-art architecture, it led to further improvement and new state-of-the-art results.

Zhong et al. \cite{b30} proposed a new augmentation technique called random erasing, where a random rectangular portion of the image is selected from the image and is set to random values, hence effectively erased. Upon experimentation, they found that this technique leads to a significant improvement in the baseline results and can be used together with other data augmentation techniques.

\section{Methodology}
In our study, we propose a variety of novel data augmentation schemes that lead to an increased ability of neural networks to learn from these datasets. These data augmentation techniques aim to mimic real-world scenarios in which concepts such as occlusion, unwanted information inclusion, etc., occur. As part of the work, we propose three sets of data augmentation techniques. The first set of data augmentation techniques involves pairwise channel transfer where we transfer the channel(R/G/B) of random image in the dataset into different images in the dataset. The second set of image augmentation techniques involves an approach where objects in an image are occluded by random images from the dataset. The final and third set of image augmentation techniques involves a novel masking approach where the images and their content are masked by horizontal, vertical, checkered, and circular stripes. We will discuss each of them in detail.

\subsection{Pairwise channel transfer}
In pairwise channel transfer, for each image in the dataset (target), we pick a random image in the dataset (source). Now, we transfer the contents of the red color channel of the source image into the red color channel of the target image by replacing the red channel of the target with the red color channel of the source image. Same can be done with green and blue color channel. Mathematically, this whole process can be represented through equation 1.
\begin{equation}
\begin{aligned}
\text{target}[T_r, T_b, T_g] = \{ & \text{target}[S_r, T_b, T_g] \\
& \leftarrow \text{source}[S_r, S_b, S_g] \}
\end{aligned}
\end{equation}
where 
\[
\begin{aligned}
    S_r & : \text{Red channel of source image}, \\
    S_g & : \text{Green channel of source image}, \\
    S_b & : \text{Blue channel of source image}, \\
    T_r & : \text{Red channel of target image}, \\
    T_g & : \text{Green channel of target image}, \\
    T_b & : \text{Blue channel of target image}.
\end{aligned}
\]

The idea behind pairwise channel transfer is to mimic the real-world phenomenon where everyday objects can be seen in different contexts with varying backgrounds. For example, a cup may be present on the dining table as well as the kitchen sink. We encounter different contexts for the cup where the background changes, but the object of interest, in this case, the cup, remains the same. The entire process of pairwise channel transfer can be understood from Figure x, which shows the complete schematic of pairwise channel transfer. Hue represents the type of color, represented in degrees on a color wheel. Saturation represents the intensity of the color, determining how vivid or washed out the color appears. Value, also known as brightness or lightness, represents the brightness of the color.

To contextualize the contents of an image, we convert each image in the dataset into the HSV color space. Afterward, we perform pairwise channel transfer, where the channel being transferred between an image in the dataset and a random image from the dataset is either Hue, Saturation, or Value (brightness). In this case, equation 1 changes into equation 2.
\begin{equation}
\begin{aligned}
\text{target}[T_h, T_s, T_v] = \{ & \text{target}[S_h, T_s, T_v] \\
& \leftarrow \text{source}[S_h, S_s, S_v] \}
\end{aligned}
\end{equation}
where 
\[
\begin{aligned}
    S_H & : \text{Hue channel of source image}, \\
    S_s & : \text{Saturation channel of source image}, \\
    S_v & : \text{Value/Brightness channel of source image}, \\
    T_h & : \text{Hue channel of target image}, \\
    T_s & : \text{Saturation channel of target image}, \\
    T_v & : \text{Value/Brightness channel of target image}.
\end{aligned}
\]

\begin{figure}
    \centering
    \includegraphics[width=\linewidth]{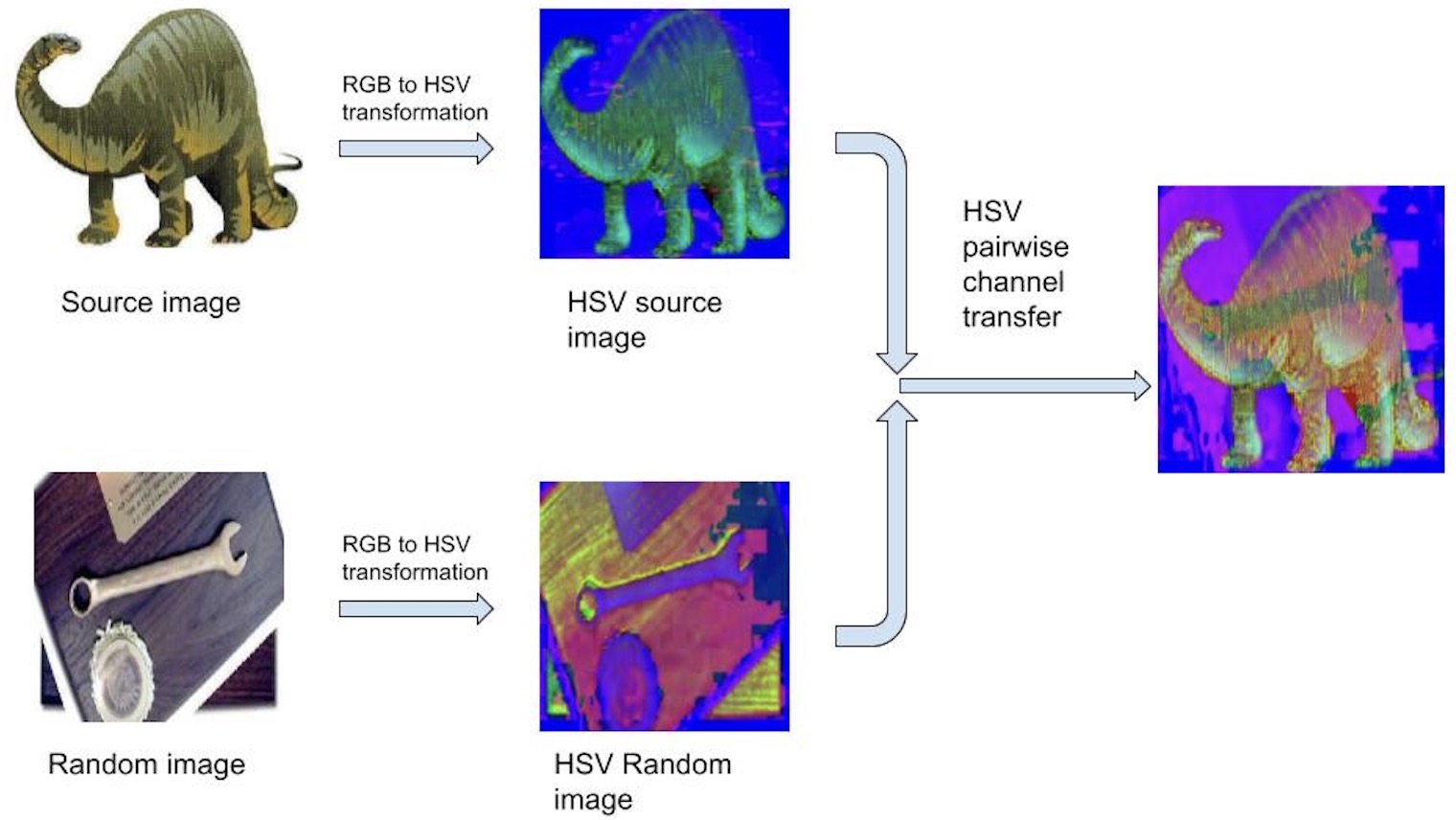} 
    \caption{Schematic diagram of pairwise channel transfer for (HSV) showing the process.}
    \label{fig:pairwise_channel_transfer}
\end{figure}

\subsection{Random Object Occlusion}
\begin{figure}
    \centering
    \includegraphics[width=\linewidth]{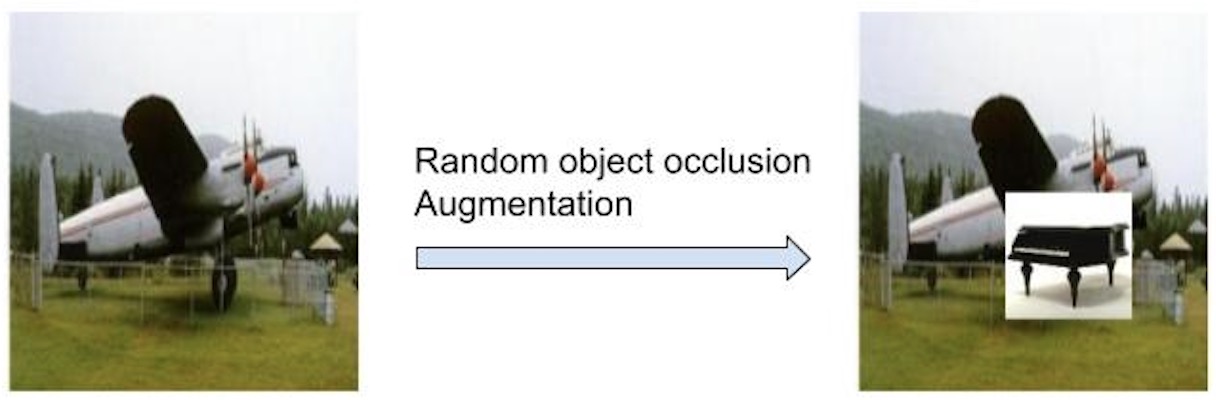} %  "occlusion.jpg" is the actual filename or path of your figure
    \caption{Schematic diagram showing the Random Object Occlusion process.}
    \label{fig:random_object_occlusion}
\end{figure}

Random object occlusion involves introduction of random objects as the occlusion in image where the random objects are the random images from the dataset. We performed this augmentation in three steps. First step is to randomly pick an image from the dataset which would act as a random object. Second step would be to resize these random objects into 100x100 resolution. Final step would be to insert this small random object into the image from the dataset, at random location, giving us the final image. The whole Random Object Occlusion augmentation can be clearly inferred from Figure \ref{fig:random_object_occlusion}.

The idea behind random object occlusion is to replicate the real world scenario where objects of interest can be occluded by random objects. For example, a person can be standing in front of the car where we are interested in detection of car. 

\subsection{Novel masking augmentaton}
\begin{figure}
    \centering
\vspace{3mm}
    \begin{subfigure}{\linewidth}
        \centering
        \includegraphics[width=\linewidth]{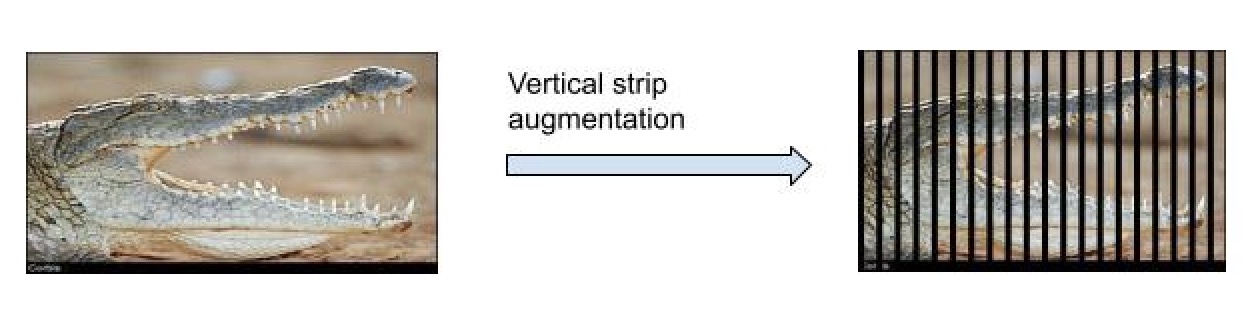}
        \caption{Vertical strip augmentation image (before and after).}
        \vspace{3mm}
        \label{fig:vertical_strip_augmentation}
    \end{subfigure}
    
    \begin{subfigure}{\linewidth}
        \centering
        \includegraphics[width=\linewidth]{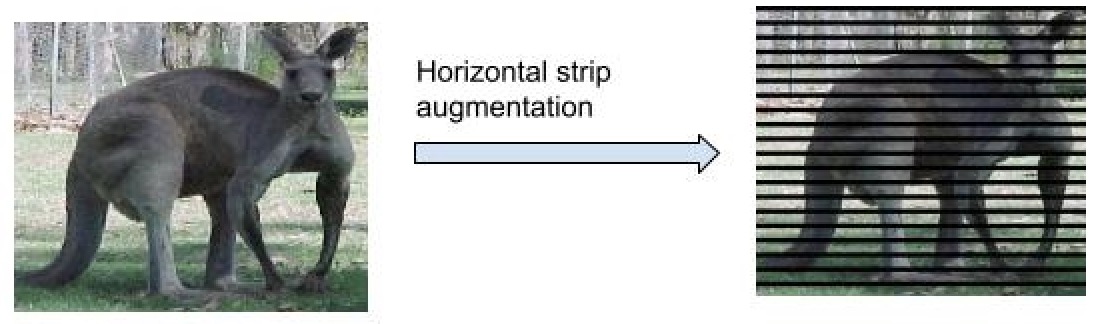}
        \caption{Horizontal strip augmentation image (before and after).}
        \vspace{3mm}
        \label{fig:horizontal_strip_augmentation}
    \end{subfigure}
    
    \begin{subfigure}{\linewidth}
        \centering
        \includegraphics[width=\linewidth]{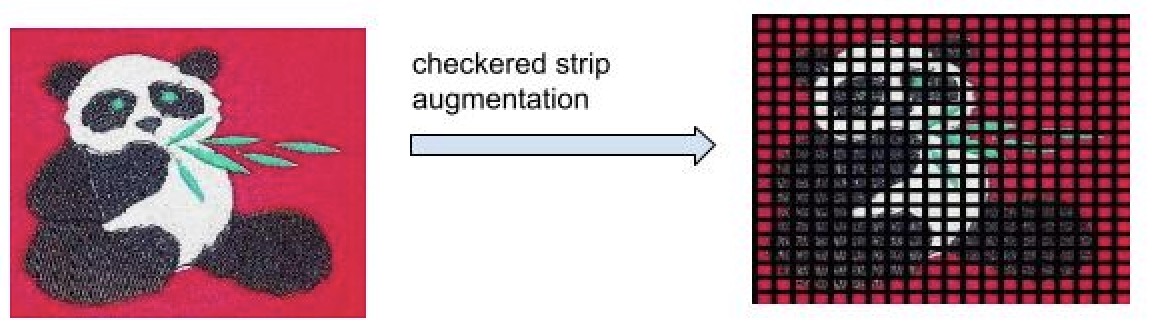}
        \caption{Checkered strip augmentation image (before and after).}
        \vspace{3mm}
        \label{fig:checkered_strip_augmentation}
    \end{subfigure}

    \begin{subfigure}{\linewidth}
        \centering
        \includegraphics[width=\linewidth]{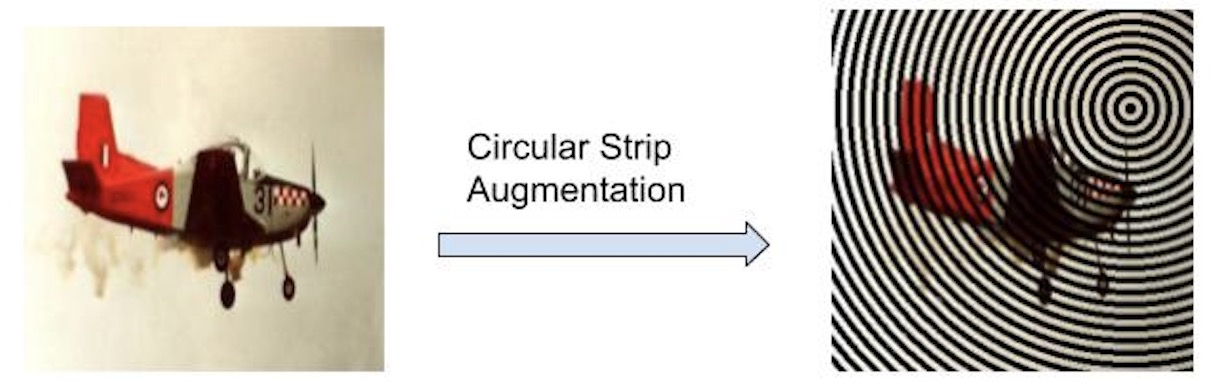}
        \caption{Circular strip augmentation image (before and after).}
        \vspace{3mm}
        \label{fig:circular_strip_augmentation}
    \end{subfigure}

    \caption{Schematic diagram showing horizontal, vertical, checkered, and circular stripes augmentation.}
    \label{fig:strip_augmentation}
\end{figure}

Novel masking augmentation involves masking the images in the dataset in new and unique ways. These novel masking approaches has been inspired by the works of Chen et. el. \cite{b26} where they use a grid to mask the content of the images leading to state of the art results on standard dataset. We propose horizontal, vertical, checkered and circular stripes as the mask for the content of the image. The result of the augmentation can be seen in contrastive display of pre and post augmentation in  Figure \ref{fig:vertical_strip_augmentation}. 

\begin{figure}
    \centering
\vspace{3mm}
    \begin{subfigure}{0.45\textwidth}
        \centering
        \includegraphics[width=\textwidth]{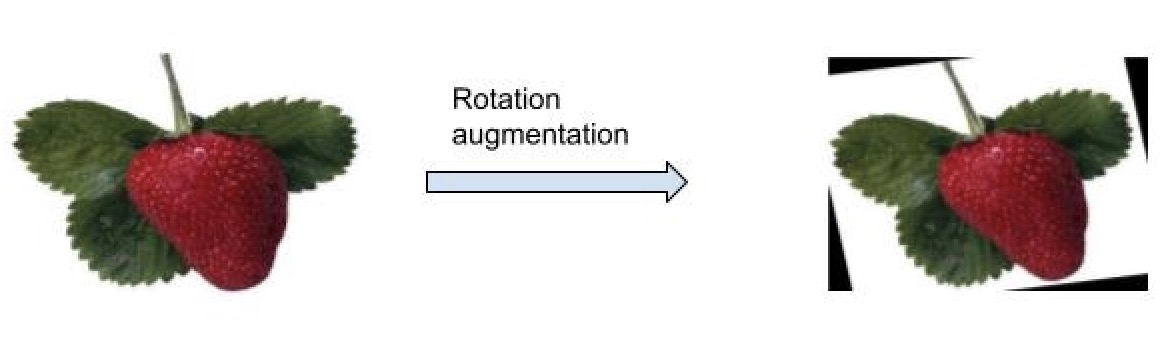}
        \caption{Rotation image (before and after).}
        \label{fig:rotation}
    \end{subfigure}
\vspace{3mm}
    \begin{subfigure}{0.45\textwidth}
        \centering
        \includegraphics[width=\textwidth]{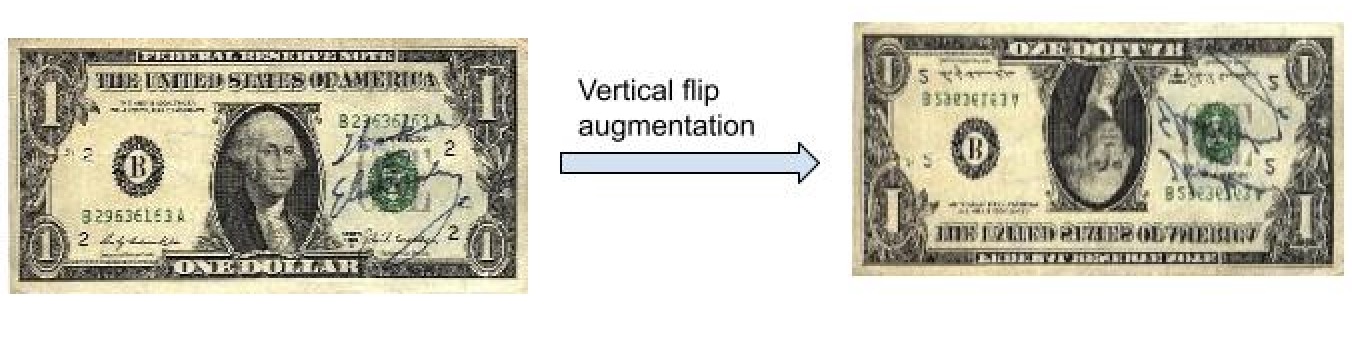}
        \caption{Vertical flip image (before and after).}
        \label{fig:vertical_flip}
    \end{subfigure}
\vspace{3mm}
    \begin{subfigure}{0.45\textwidth}
        \centering
        \includegraphics[width=\textwidth]{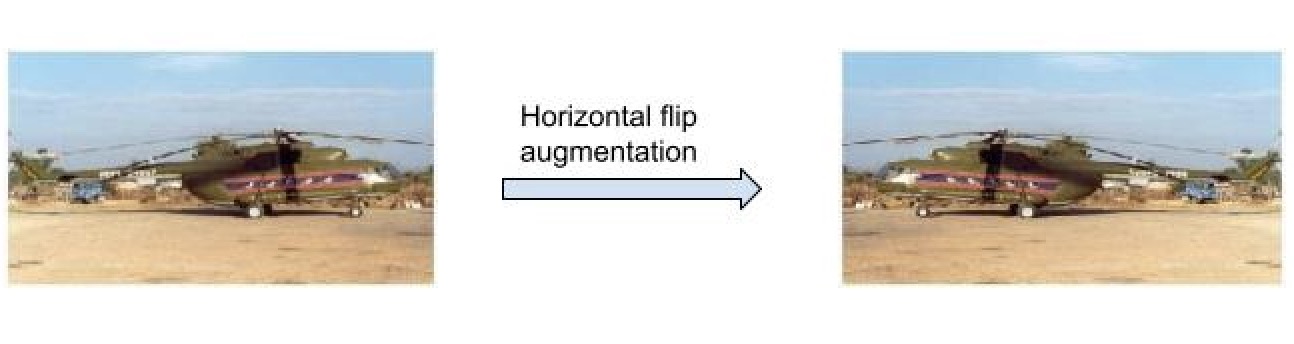}
        \caption{Horizontal flip image (before and after).}
        \label{fig:horizontal_flip}
    \end{subfigure}
    \vspace{3mm}
    \begin{subfigure}{0.45\textwidth}
        \centering
        \includegraphics[width=\textwidth]{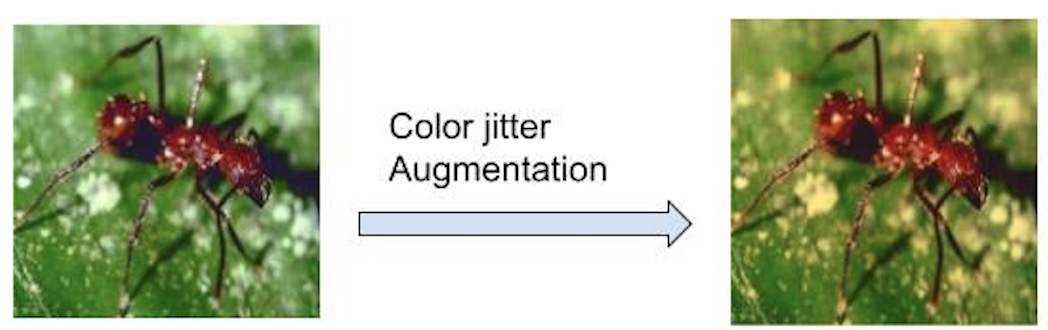}
        \caption{Color jitter image (before and after).}
        \label{fig:color_jitter}
    \end{subfigure}
\vspace{3mm}
    \begin{subfigure}{0.45\textwidth}
        \centering
        \includegraphics[width=\textwidth]{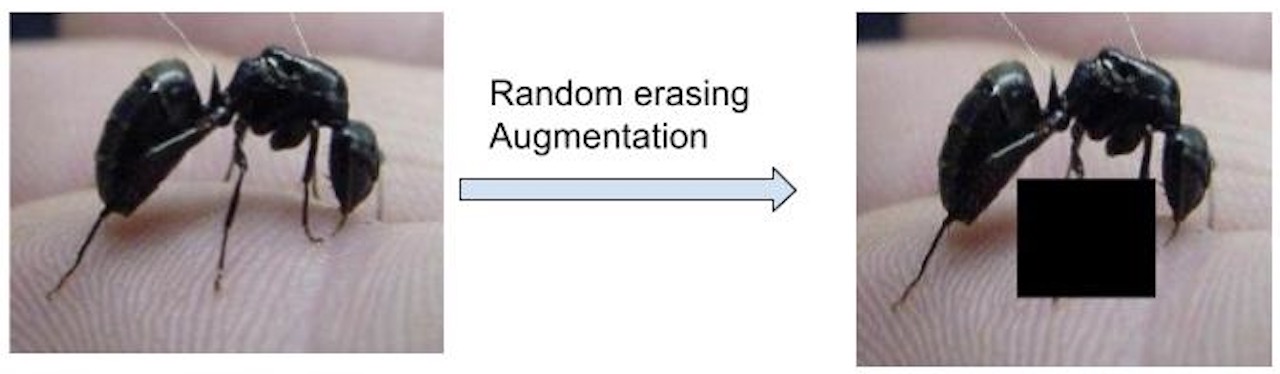}
        \caption{Random erasing image (before and after).}
        \label{fig:random_erasing}
    \end{subfigure}
    \caption{Schematic diagram showing various image augmentation techniques.}
    \label{fig:existing_image_augmentation}
\end{figure}

To ascertain the qualitative improvements in the results due to the newly proposed data augmentation technique in this work, we compare our proposed data augmentation technique with the pre-existing data augmentation technique. The pre-existing data augmentation techniques that we benchmark our data augmentation technique against are Rotation, Horizontal and Vertical Flips, Resizing, Translation, Blur, Color Jitter, and Random Erasing. These augmentation techniques are self-explanatory. During Rotation augmentation, we rotate the image by a random angle between 0 and 45 degrees. During Horizontal and Vertical flip augmentation, we flip the image horizontally and vertically, respectively. Resizing as an augmentation leads to a change in the resolution of the image. In this work, we resized all images in the dataset to 300x300. Translation involves moving the contents of the image horizontally as well as vertically. For blur augmentation, we used Gaussian blur as the kernel to generate the final augmented image. Color Jitter involves changing the Red, Green, and Blue color channels by random values, which leads to a random change in the color of the image. Random Erasing, which is a widely used augmentation technique, involves erasing random parts of the image. All these pre-existing augmentation techniques and the changes that they introduce in the image can be understood from Figure \ref{fig:existing_image_augmentation}. 

\begin{figure*}
    \centering
    \includegraphics[width=\textwidth]{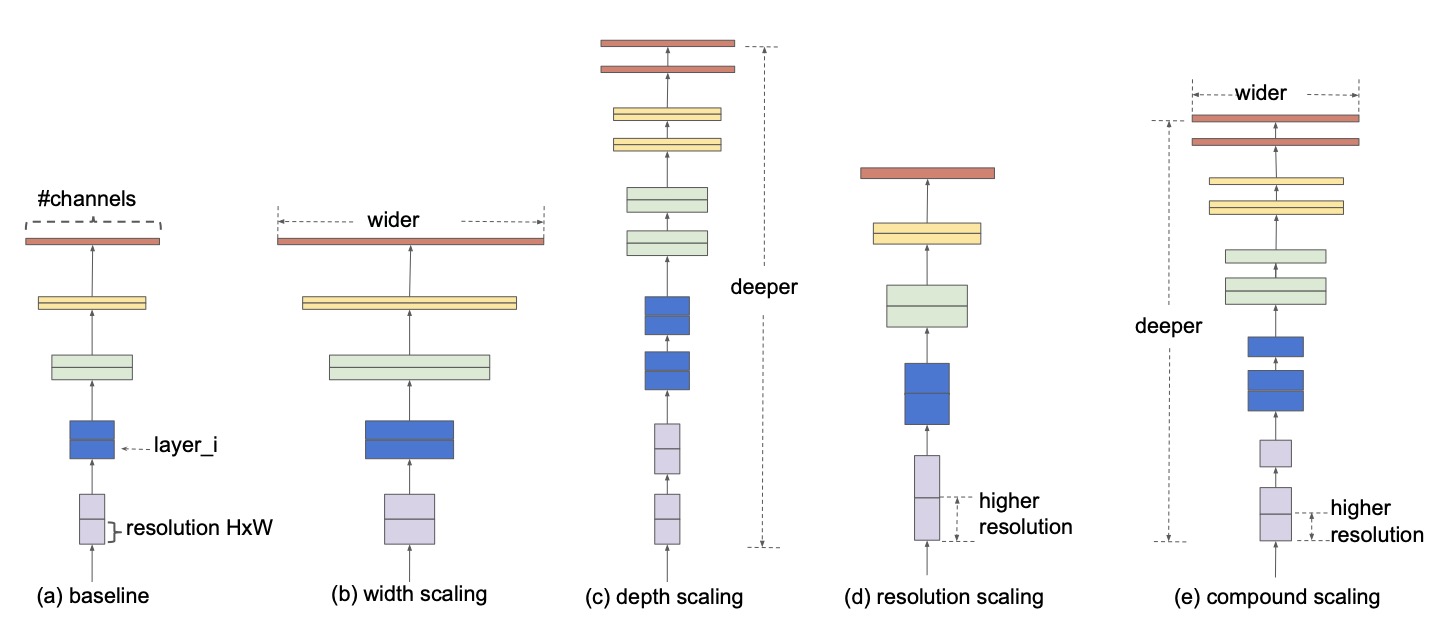}
    \caption{Variations of different EfficientNet models. \cite{b16}}
    \label{fig:efficientnet_variations}
\end{figure*}

The Deep Learning model that we use to evaluate our work is called EfficientNet, which has multiple versions differing from each other in terms of depth, width, and resolution, and thus the number of trainable parameters in the network. These different variants and the differences between them can be well understood from Figure \ref{fig:efficientnet_variations}, which has been borrowed from \cite{b16}.

\section{Experimentation}
We performed a series of experiments to ascertain the performance of the proposed data augmentation technique. Our experiments involved the Caltech-101 dataset as the source of images. This dataset has a total of 9,146 images belonging to 101 classes. These images vary in resolution. To benchmark the proposed data augmentation technique, we used EfficientNet\_b0 \cite{b16} as the deep learning model. We utilized Python and PyTorch to train the model and Matplotlib Python packages to log the results. To train this deep learning model, we used cloud computing in the form of Google Colab Pro+ to fulfill our computing resources. This cloud computing platform offers GPUs such as Nvidia A100, Nvidia V100, and Nvidia T4. We utilized the Nvidia A100 GPU to train the aforementioned deep learning model.

The first step of the experimentation was dataset preparation. For each proposed image augmentation technique, we created a separate copy of the dataset, which is the augmented version of the dataset. This led to the creation of seven augmented versions of the dataset. Therefore, in total, we had 8 versions of the dataset. All these 8 versions of the dataset are then merged together to form a single augmented dataset. This final augmented dataset has a total of 82,281 images. This dataset contains all novel proposed data augmentation techniques along with the original dataset.

We also applied the existing augmentation techniques that we have discussed above to each image in the dataset. The number of existing augmentation techniques that we investigated is six. Therefore, we created six different augmented versions of the dataset. Afterwards, we classwise merged all these different augmented datasets to form a single augmented dataset containing all existing augmentations.

With all this data wrangling and preprocessing, we have three different datasets. The first dataset is the original dataset which contains images without any augmentation. The second dataset contains all the original images along with augmented images with the newly proposed data augmentation techniques. The third dataset contains all the original images along with augmented images with the existing image augmentation techniques.

\begin{flushleft}
\begin{description}
    \item[Dataset 1] \hspace{1em}: Original dataset without any augmentation
    \item[Dataset 2] \hspace{0.9em} : Dataset containing all novel augmentation dataset
    \item[Dataset 3] \hspace{0.9em} : Dataset contain ing all existing augmentation dataset
\end{description}
\end{flushleft}

Corresponding to three different datasets, we trained the EfficientNet\_b0 model on these three different datasets for fifty epochs each. While training these models, we used a learning rate of 0.001 and a batch size of 64. The batch size was chosen to fit in the VRAM without running out of memory. We used a training-validation split of 0.8:0.2 during model training.

While training the model, we observed for overfitting based on the trend line in training and validation error. Overfitted models don't generalize well to real-world unseen examples. Upon training, we determined the accuracy without overfitting, which gives a real-world performance representation. We obtained the accuracy metrics as well as training and validation errors for each epoch of training. We discuss all the experimentation results in the next section.

\section{Results}
\label{sec:results}
\begin{figure}
    \centering
    \includegraphics[width=\columnwidth]{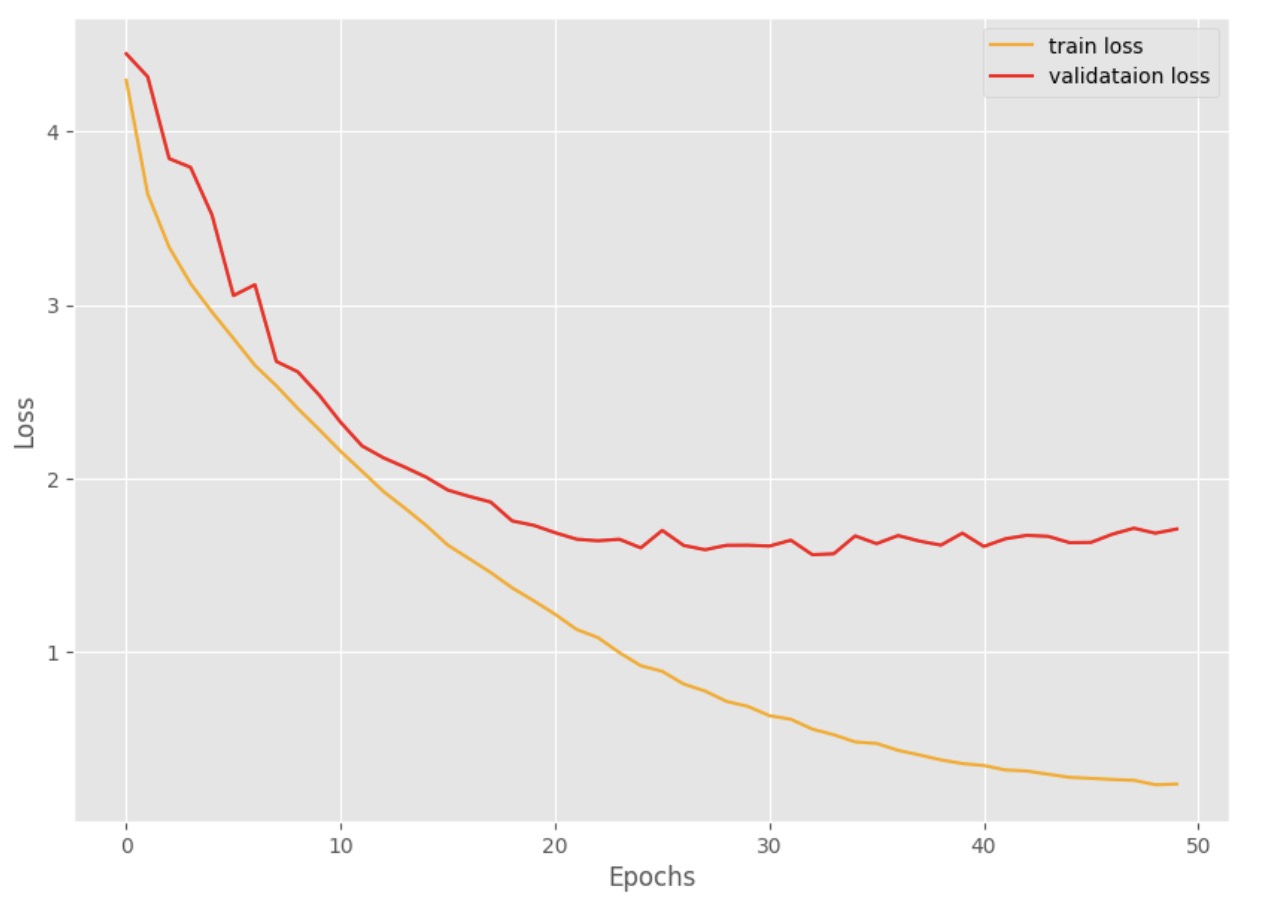}
    \caption{Training Loss for Dataset 1. As we can see the overfitting starts around 10th epoch as it's evident from diverging training and validation loss.}
    \label{fig:training_loss_dataset1}
\end{figure}

\begin{figure}
    \centering
    \includegraphics[width=\columnwidth]{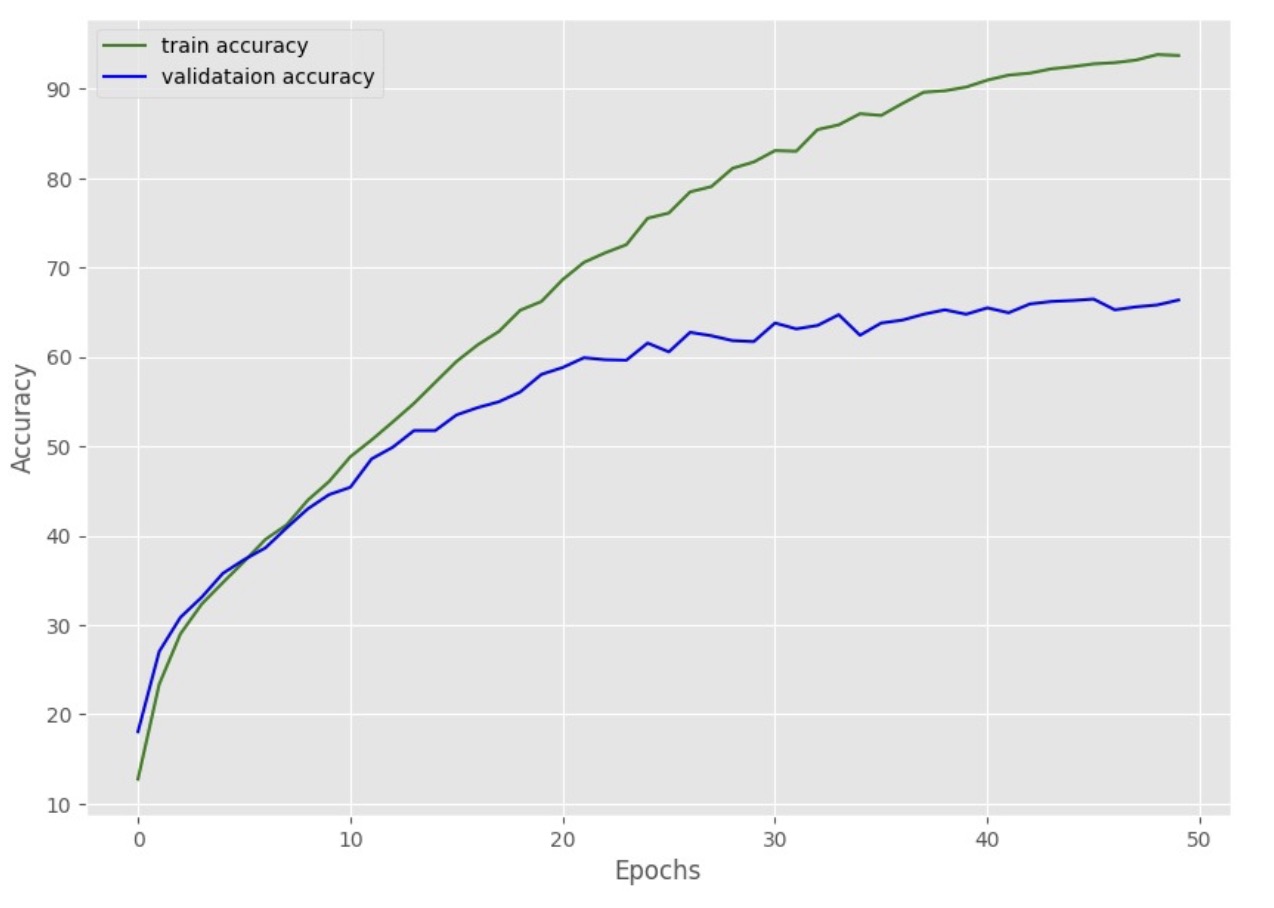}
    \caption{Accuracy for Dataset 1. As can be seen the accuracy metrics starts diverging at 10th epoch which shows the model is overfitting after 10th epoch.}
    \label{fig:accuracy_dataset1}
\end{figure}

 We trained EfficioentNet\_b0 model model on Dataset 1, Dataset 2 and Dataset 3 separately.  Figure \ref{fig:training_loss_dataset1} shows the training and validation loss as a function of number of Epochs when we trained the model on Dataset 1. Figure \ref{fig:accuracy_dataset1} shows the accuracy as a function of Epochs when the EfficioentNet\_b0 model is trained on Dataset 1. As we can see from Figure \ref{fig:training_loss_dataset1}, the model starts to overfit around the 12th epoch. Around the 12th epoch, the training loss and validation loss starts to diverge from each other which indicates overfitting.

\begin{figure}
    \centering
    \includegraphics[width=\columnwidth]{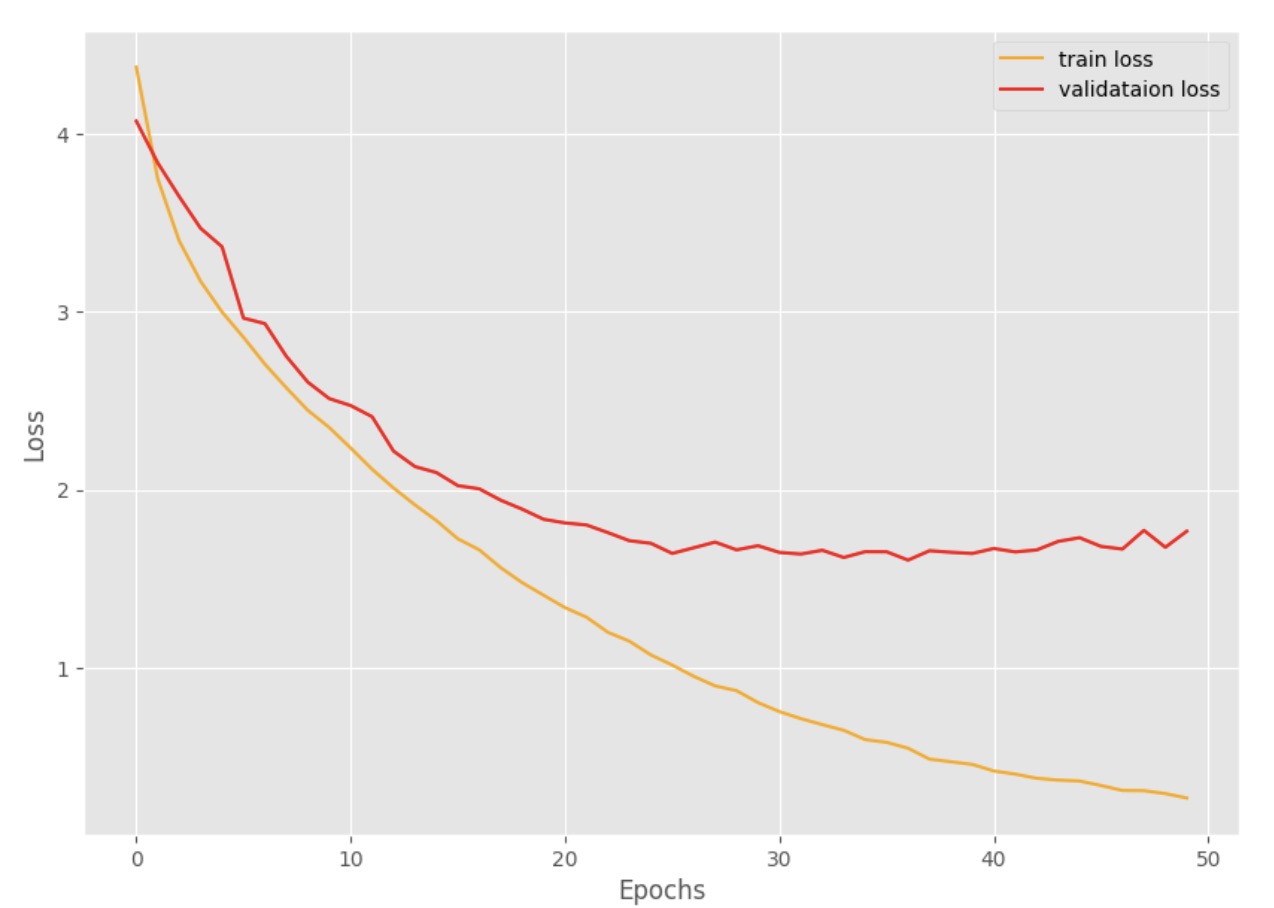}
    \caption{Training Loss and validation loss for Horizontal Strip Augmentation. They start to diverge around 12th epoch which implies overfitting.}
    \label{fig:train_loss_horizontal_strip}
\end{figure}

\begin{figure}
    \centering
    \includegraphics[width=\columnwidth]{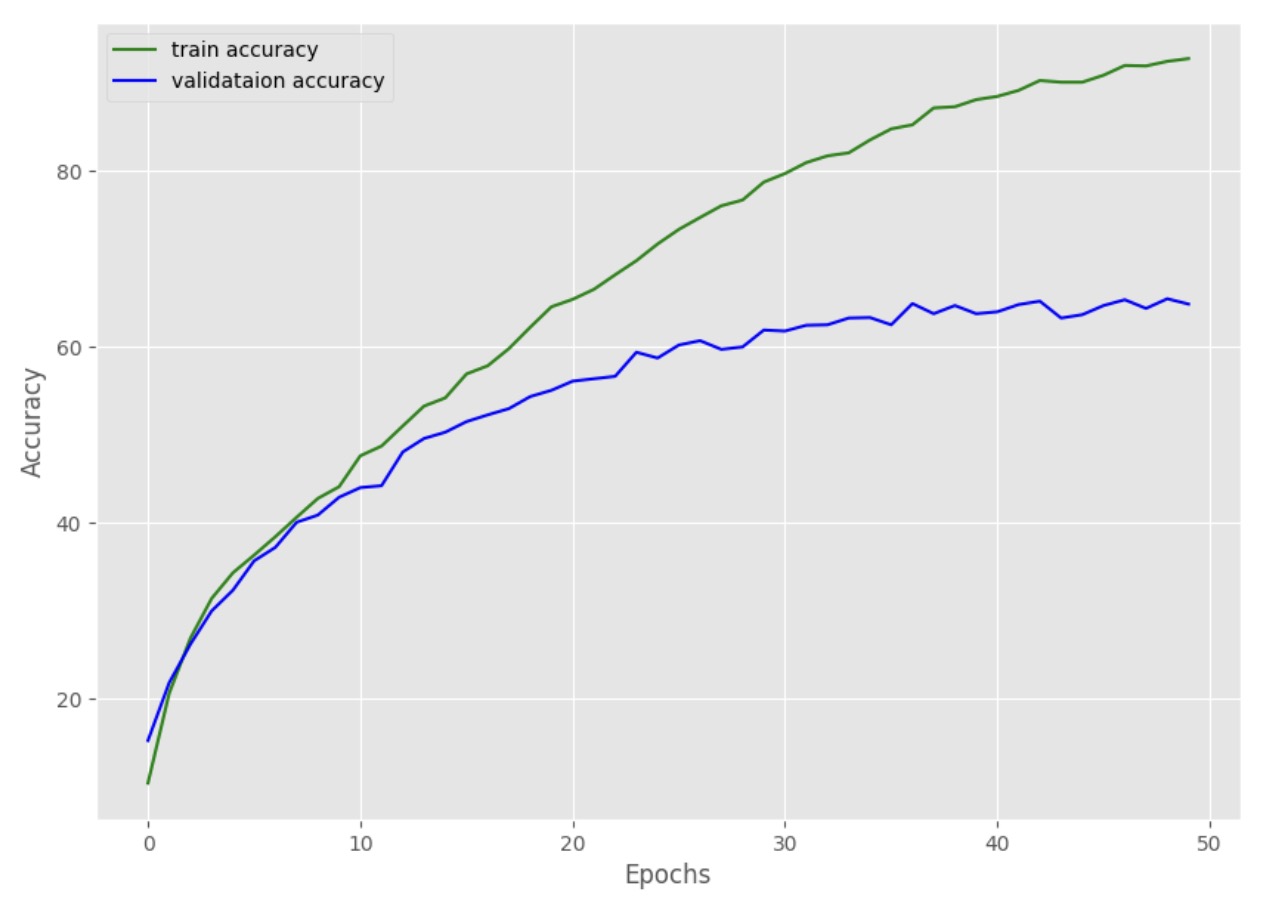}
    \caption{Accuracy for Horizontal Strip Augmentation. As is clear from the figure, overfitting starts around 12th epoch.}
    \label{fig:accuracy_horizontal_strip}
\end{figure}

 We tried to understand the effect of individual augmentation on the accuracy. Therefore we trained individually augmented dataset with different newly proposed  augmentation technique on the EfficientNet\_B0 model. Figure \ref{fig:train_loss_horizontal_strip} shows the training and validation loss as a function of epochs for horizontal strip augmentation. Figure \ref{fig:accuracy_horizontal_strip} shows the accuracy as a function of epochs for horizontal strip augmentation. As we can see the there is slight delay in onset of divergence between training and validation loss when compared to the vanilla dataset without any augmentation. This indicates that the model learns a slightly better representation from the dataset. 

\begin{figure}
    \centering
    \includegraphics[width=\columnwidth]{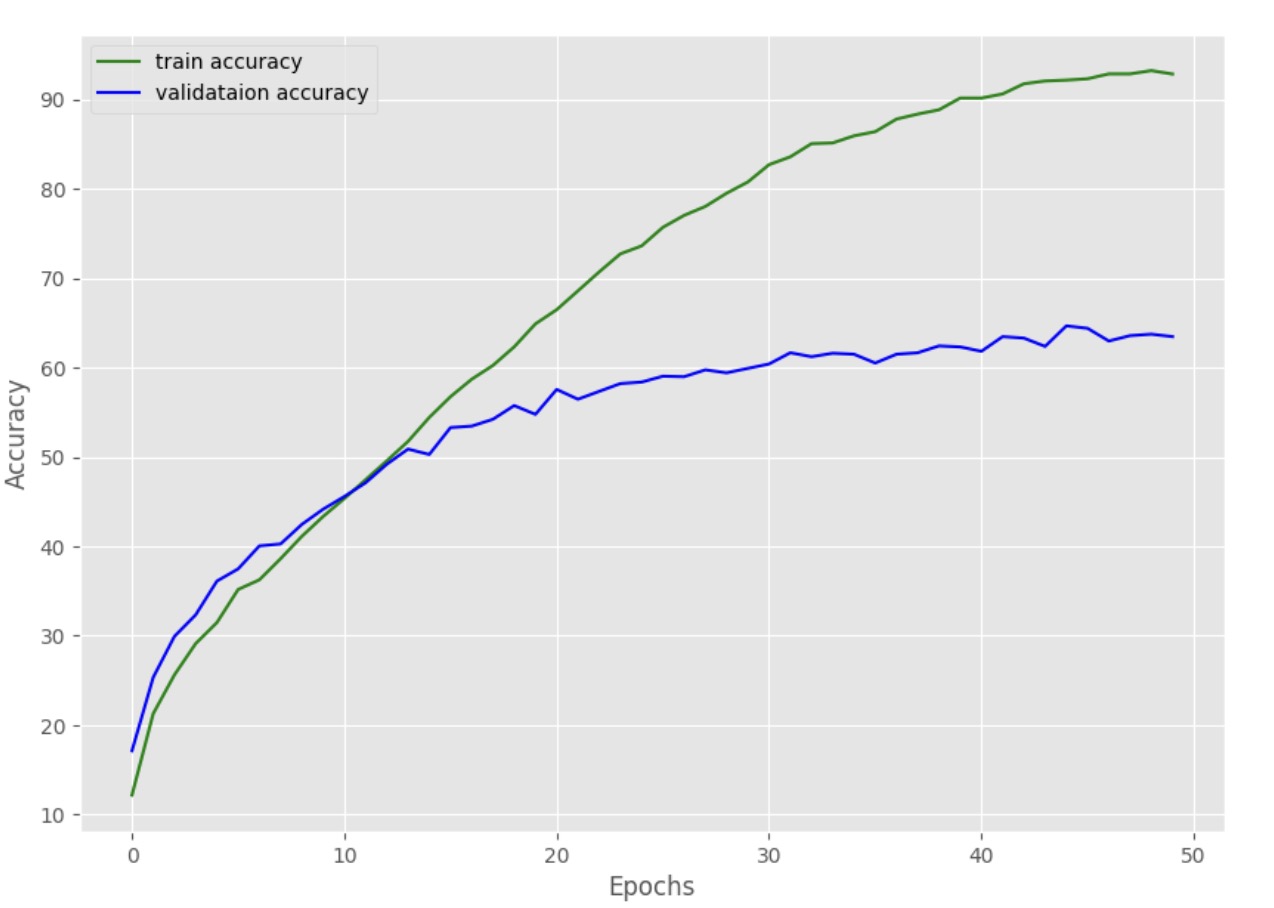}
    \caption{Training and validation accuracy for Hue Saturation Channel Transfer. Training and validation accuracy diverges around 15th epoch. This implies a slight delay in overfitting compared to unaugmented dataset training in Fig \ref{fig:accuracy_dataset1}}
    \label{fig:hue_saturation_channel_transfer}
\end{figure}

Similarly, we tried to study the effect of the Hue, Saturation  pairwise channel transfer augmentation technique. Figure \ref{fig:hue_saturation_channel_transfer} contains the accuracy value as a function of epoch for hue, saturation pairwise channel transfer augmentation. As we can see from the figure the onset of overfitting is delayed due to augmentation. This leads to an improved accuracy without overfitting. 

\begin{figure}
    \centering
    \includegraphics[width=\columnwidth]{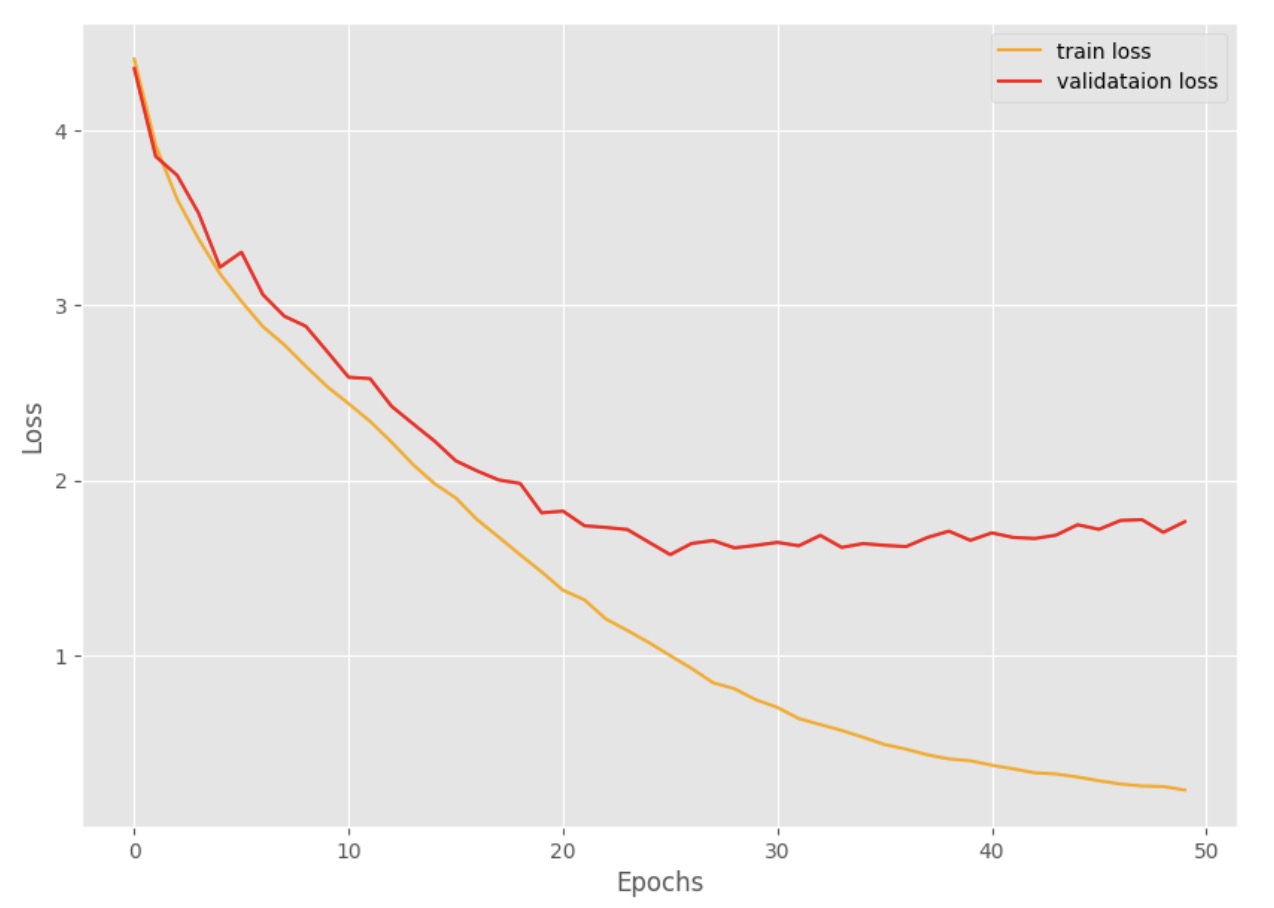}
    \caption{Training and validation loss for Pairwise Channel Transfer Augmentation. 18th epoch markes the point of overfitting as training and validation loss start to diverge.}
    \label{fig:training_pairwise_channel_transfer}
\end{figure}

\begin{figure}
    \centering
    \includegraphics[width=\columnwidth]{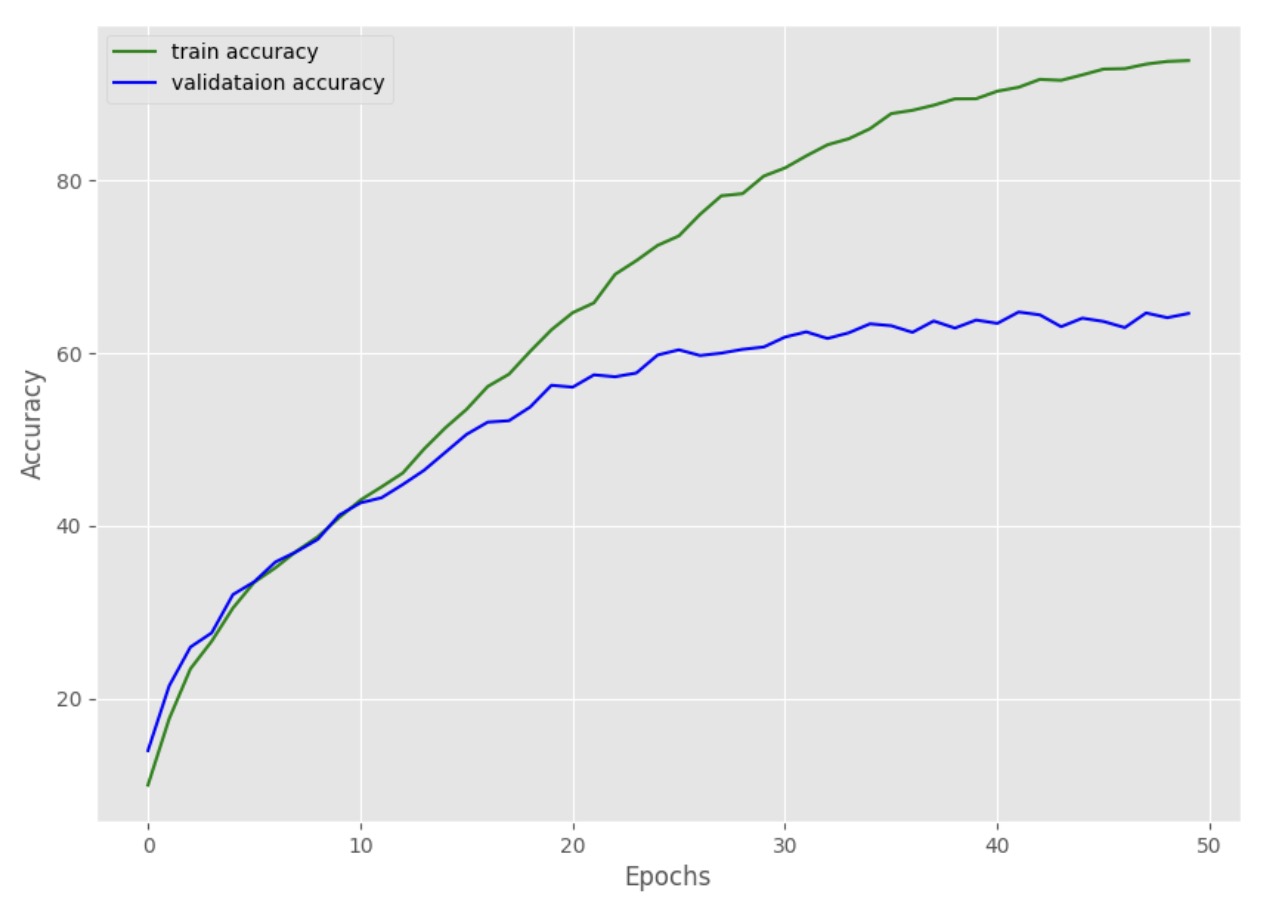}
    \caption{Accuracy for Pairwise Channel Transfer augmentaton. Accuracy value for training and validation phase starts to diverge around 18th epoch indicating overfitting.}
    \label{fig:accuracy_pairwise_channel_transfer}
\end{figure}

As we did for horizontal strip augmentation, we studied pairwise channel transfer  and Fig \ref{fig:training_pairwise_channel_transfer} contains the training loss and validation loss for pairwise channel transfer augmentation as a function of epoch. Also, Figure \ref{fig:accuracy_pairwise_channel_transfer} contains the accuracy for pairwise channel transfer augmentation as a function of epoch. In a predictable fashion, we can see that the onset of overfitting is delayed which leads to an increase in accuracy. 

\begin{figure}
    \centering
    \includegraphics[width=\columnwidth]{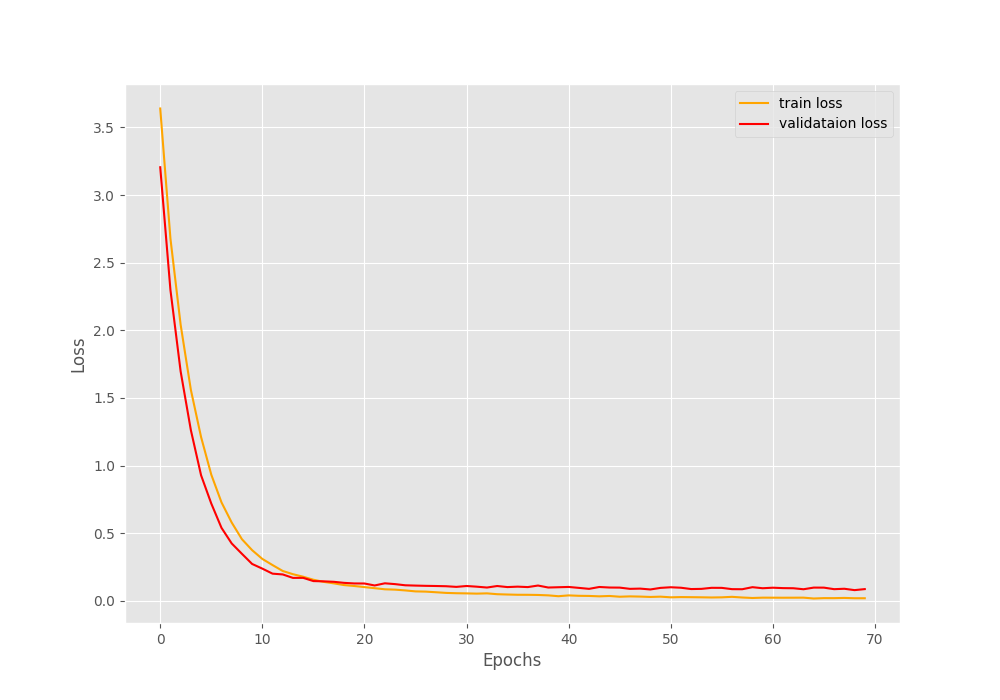}
    \caption{Training and Validation Loss for Dataset 2 (All new image augmentation techniques are used)}
    \label{fig:training_loss_dataset_2}
\end{figure}

\begin{figure}
    \centering
    \includegraphics[width=\columnwidth]{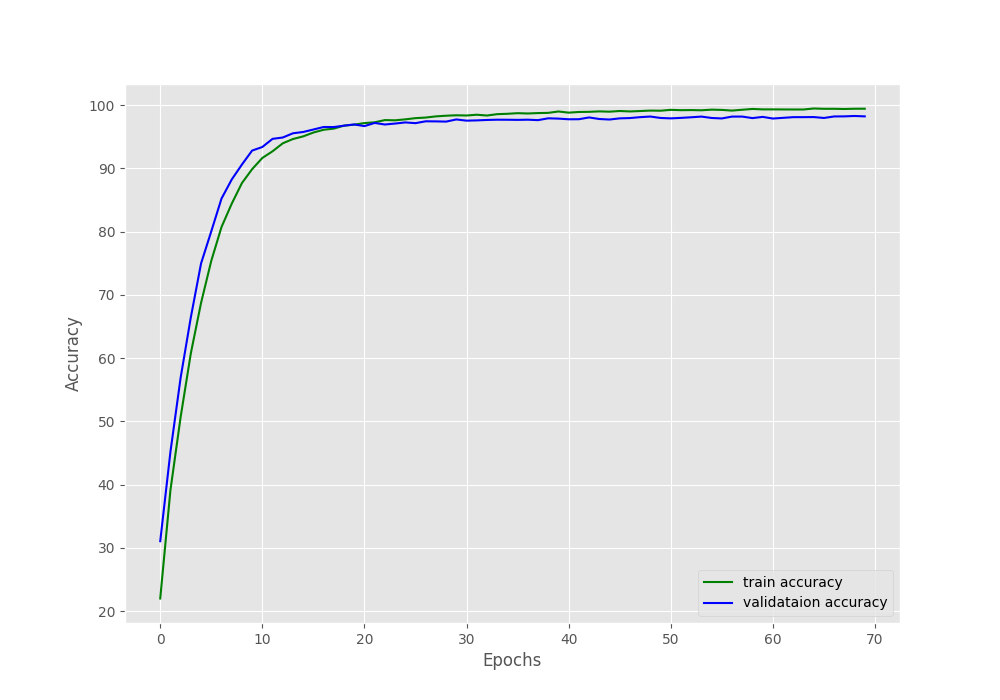}
    \caption{Training and Validation Accuracy for Dataset 2 (All new image augmentation techniques are used.)}
    \label{fig:accuracy_dataset_2}
\end{figure}

The fully augmented dataset, which incorporates images from all different proposed augmentation techniques, forms Dataset 2. We trained the model for 50 epochs. The training and validation losses for this training run can be found in Fig \ref{fig:training_loss_dataset_2}, and the training and validation accuracies are detailed in Fig \ref{fig:accuracy_dataset_2}. As we can observe, there is no sign of overfitting until the 19th epoch, and the accuracy at the onset of overfitting reaches 96.74\%. Furthermore, both the training loss and validation loss do not diverge drastically, unlike the case with individual image augmentation techniques, implying that overfitting is less severe in this case.

\begin{figure}
    \centering
    \includegraphics[width=\columnwidth]{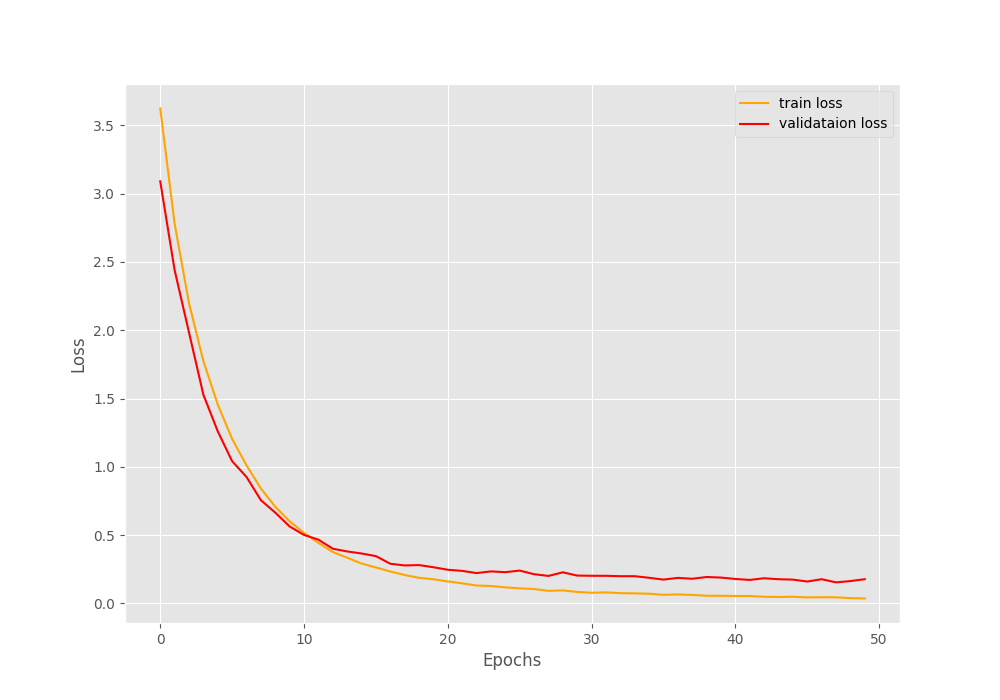}
    \caption{Training and Validation Accuracy for Dataset 3 (Only existing image augmentation techniques are used)}
    \label{fig:train_dataset_3}
\end{figure}

\begin{figure}
    \centering
    \includegraphics[width=\columnwidth]{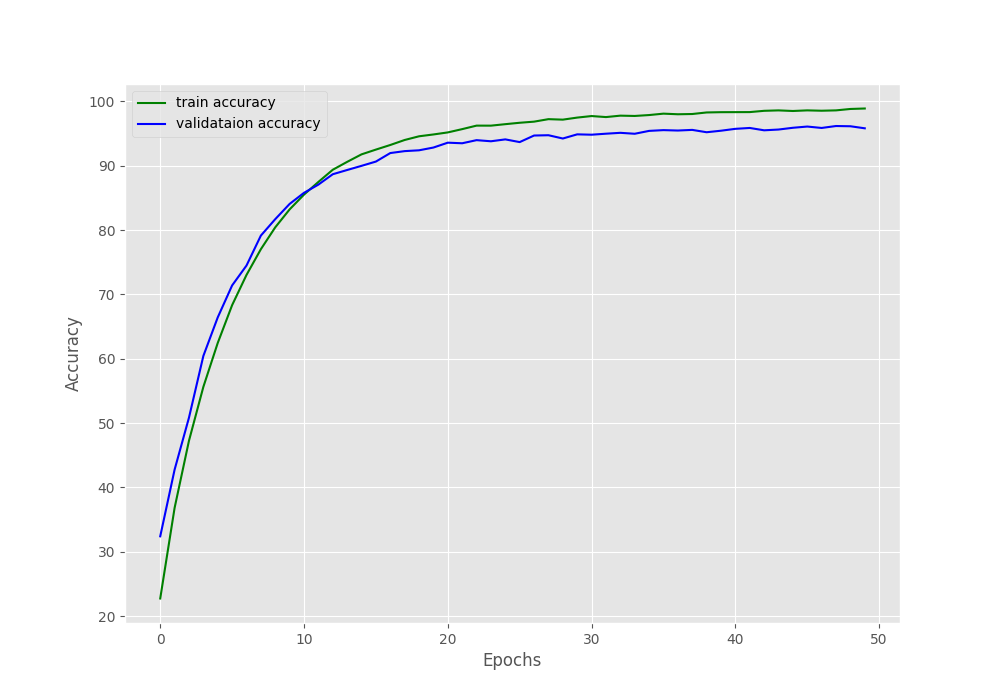}
    \caption{Training and Validation loss for Dataset 3 (only existing image augmentation techniques are used.)}
    \label{fig:accuracy_dataset_3}
\end{figure}

A similar trend can be observed when we train the EfficientNet\_B0 model on Dataset 3, which is composed of data formed after applying existing image augmentation techniques. We train the model for 50 epochs using the settings described in Section \ref{sec:results}. The training and validation loss continue to decrease until overfitting begins at epoch 11, as illustrated in Fig \ref{fig:train_dataset_3}. The accuracy of the model reaches 85.78\% before the onset of overfitting, as shown in Fig: \ref{fig:accuracy_dataset_3}. This performance is lower than the 96.7\% accuracy achieved with our proposed novel image augmentation techniques, highlighting the effectiveness of our methods. 

Table \ref{tab:model_training_summary} gives a summary of the models and overfitting epoch around which they start to overfit. As we can see that the addition of augmentation to the dataset leads to delayed overfitting and thus better learning by the model. 

\begin{table*}
    \centering
    \begin{tabular*}{\textwidth}{l@{\extracolsep{\fill}}cccc}
        \toprule
        Model name & Augmentation technique & Epochs (trained for) & Overfitting epoch & Accuracy (before overfitting) \\
        \midrule
        EfficientNet\_b0 & No augmentation & 50 & 10 & 44.0 \\
        EfficientNet\_b0 & Horizontal strip augmentation & 50 & 12 & 44.20 \\
        EfficientNet\_b0 & Hue Saturation Channel transfer & 50 & 15 & 50.27 \\
        EfficientNet\_b0 & Pairwise channel transfer & 50 & 18 & 52.13 \\
        EfficientNet\_b0 & \textbf{All proposed augmentations combined} & \textbf{50} & \textbf{19} & \textbf{96.740} \\
        EfficientNet\_b0 & All existing augmentations combined & 50 & 11 & 85.782 \\
        \bottomrule
    \end{tabular*}
    \caption{Summary of model training with different augmentation techniques.}
    \label{tab:model_training_summary}
\end{table*}

\section{Discussion}
Data augmentation is a critical aspect of Machine Learning and Deep Learning. The use of data augmentation contributes to improved learning by models, thus serving as a reliable method for achieving better results while keeping other parameters constant. During data augmentation, various characteristics of an image are manipulated to simulate real-world variations. For instance, augmenting brightness aims to capture the variability in lighting found in the real world. Similarly, rotation seeks to represent the orientations of subjects of interest, mirroring how objects are encountered in daily life. Employing such data augmentation techniques ensures that the dataset reflects the diverse conditions found in real-world scenarios. In general, incorporating a broader range of unique data augmentation techniques leads to enhanced model performance, as evidenced by our study.

n our investigation, we observed that without any augmentation, the model's performance was subpar, with an accuracy of 44\% \ref{tab:model_training_summary}. Training the model using individually augmented datasets revealed that each technique independently resulted in a slight improvement in the model's accuracy. When combining all augmentation techniques, the improvements from individual augmentations seemed to have a cumulative effect, resulting in an overall accuracy of 96.7\% \ref{tab:model_training_summary}. When compared to existing image augmentation techniques, which yield an accuracy of 87.78\%, we observed an increase of 8.92\% in the accuracy of the same model. This suggests that the newly proposed image augmentation techniques significantly enhance the model's ability to learn new patterns in the data without overfitting. This overfitting, in the training plots, is characterized by the point where the training loss overshoots the validation loss.

Furthermore, we noted a delay in the onset of overfitting in terms of the number of epochs due to the use of data augmentation techniques. In our opinion, the delays induced by individual augmentation techniques had a cumulative effect, collectively contributing to reduced overfitting in the model.

\section{Conclusion}
In this work, we propose three new data augmentation techniques. The first technique, pairwise channel transfer, involves forming pairs of two images: one selected sequentially from the dataset and another chosen randomly from the same dataset (see Figs. \ref{fig:pairwise_channel_transfer} and \ref{fig:hue_saturation_channel_transfer}). We then transfer the individual channels from the randomly chosen image to the corresponding channels of the first image. This process facilitates the cross-pollination of information between the two images, integrating elements of the random image into the dataset. This can be likened to placing a subject of interest into a new environment, thereby introducing novel information.

The second technique is a novel occlusion approach (see Fig \ref{fig:random_object_occlusion}): we select random images from the dataset to serve as occlusions for other images. This approach simulates scenarios where a non-interesting subject blocks the subject of interest, helping the model become occlusion-invariant.

The third data augmentation technique involves novel masking methods. In this approach, we use different types of masks to conceal the contents of the image. The masks used include vertical strips (see Fig \ref{fig:vertical_strip_augmentation}), horizontal strips (see Fig \ref{fig:horizontal_strip_augmentation}), checkered strips (see Fig \ref{fig:checkered_strip_augmentation}), and circular strips (see Fig \ref{fig:circular_strip_augmentation}). These diverse masking approaches were inspired by \cite{b26} and aim to introduce robustness to the model.

Furthermore, we compare the performance of each individual augmentation technique and its impact on accuracy. Additionally, we juxtapose the proposed augmentation techniques with existing methods such as rotation (see Fig \ref{fig:rotation}), vertical flip (see Fig \ref{fig:vertical_flip}), horizontal flip (see Fig \ref{fig:horizontal_flip}), color jitter (see Fig \ref{fig:color_jitter}), and random erasing (see Fig \ref{fig:random_erasing}).

Upon comparison, we discovered that our newly proposed data augmentation techniques led to a significant 50\% improvement in the accuracy of the EfficientNet\_B0 model. We also noted a delayed onset of overfitting when the model was trained on the augmented dataset, thereby demonstrating the efficacy of the proposed data augmentation techniques.

\section{Future Work}
In the future, we plan to conduct a fine-grained investigation of the proposed data augmentation techniques on various datasets to ascertain their widespread effectiveness. The dataset used in this paper for investigation is a generic dataset containing 101 classes from day-to-day life. However, we intend to explore the effectiveness of our proposed data augmentation techniques on datasets from different genres, such as biomedical images, hyperspectral images, and aerial images. Such a comprehensive study would determine the widespread effectiveness of the proposed data augmentation techniques.


\begin{thebibliography}{30}
    \bibitem{b1} J. Wang and L. Perez, "The Effectiveness of Data Augmentation in Image Classification using Deep Learning," Stanford University, 450 Serra Mall, Email: zwang01@stanford.edu, Dec. 2017.
    
    \bibitem{b2} N. M. R. Aquino, M. Gutoski, L. T. Hattori, and H. S. Lopes, "The Effect of Data Augmentation on the Performance of Convolutional Neural Networks," Federal University of Technology - Paraná, Av. Sete de Setembro, 3165 - Rebouças CEP 80230-901, Emails: nmarceloromero@gmail.com, matheusgutoski@gmail.com, lthattori@gmail.com, hslopes@utfpr.edu.br.
    
    \bibitem{b3} L. Pham, C. Le, D. Ngo, A. Nguyen, J. Lampert, A. Schindler, and I. McLoughlin, "A Light-weight Deep Learning Model for Remote Sensing Image Classification," Austrian Institute of Technology, Vienna, Austria, Email: lam.pham@ait.ac.at; HCM University of Technology, HCM, VietNam, Email: cam.levt123@hcmut.edu.vn; University of Essex, Colchester, UK, Email: dn22678@essex.ac.uk; FPT Soft Company, HCM, VietNam, Email: AnhNTN34@fsoft.com.vn; Singapore Institute of Technology, Singapore, Email: ian.mcloughlin@singaporetech.edu.sg, Feb. 2023.
    
    \bibitem{b4} E. D. Cubuk*, B. Zoph*, D. Mane, V. Vasudevan, and Q. V. Le, "AutoAugment: Learning Augmentation Strategies from Data," Google Brain, April 2019.
    
    \bibitem{b5} W. Li, C. Chen, M. Zhang, H. Li, and Q. Du, "Data Augmentation for Hyperspectral Image Classification With Deep CNN," IEEE Geoscience and Remote Sensing Letters, vol. 16, no. 4, pp. 593-597, April 2019.
    
    \bibitem{b6} P. Chen, S. Liu, H. Zhao, and J. Jia, "GridMask Data Augmentation," The Chinese University of Hong Kong, {pgchen, sliu, hszhao, leojia}@cse.cuhk.edu.hk, Jan. 2020.
    
    \bibitem{b7} T. DeVries and G. W. Taylor, "Improved Regularization of Convolutional Neural Networks with Cutout," University of Guelph, Canadian Institute for Advanced Research and Vector Institute, Nov. 2019.
    
    \bibitem{b8} R. C. Fong and A. Vedaldi, "Occlusions for Effective Data Augmentation in Image Classification," University of Oxford, Facebook AI Research, 2019 IEEE/CVF International Conference on Computer Vision Workshop (ICCVW).
    
    \bibitem{b9} Z. Zhong, L. Zheng, G. Kang, S. Li, and Y. Yang, "Random Erasing Data Augmentation," 1Department of Artificial Intelligence, Xiamen University, 2ReLER, University of Technology Sydney, 3Research School of Computer Science, Australian National University, 4School of Computer Science, Carnegie Mellon University, The Thirty-Fourth AAAI Conference on Artificial Intelligence (AAAI-20).
    
    \bibitem{b10} J. Wang and L. Perez, "The Effectiveness of Data Augmentation in Image Classification using Deep Learning," Stanford University, 450 Serra Mall, zwang01@stanford.edu, luis0@stanford.edu, Dec. 2017.
    
    \bibitem{b11} Y. LeCun, L. Bottou, Y. Bengio, and P. Haffner, "Gradient-Based Learning Applied to Document Recognition," Proceedings of the IEEE, November 1998.
    
    \bibitem{b12} A. Krizhevsky, I. Sutskever, and G. E. Hinton, "ImageNet Classification with Deep Convolutional Neural Networks," University of Toronto, kriz@cs.utoronto.ca, ilya@cs.utoronto.ca, hinton@cs.utoronto.ca.
    
    \bibitem{b13} A. Dosovitskiy, L. Beyer, A. Kolesnikov, D. Weissenborn, X. Zhai, T. Unterthiner, M. Dehghani, M. Minderer, G. Heigold, S. Gelly, J. Uszkoreit, and N. Houlsby, "An Image is Worth 16x16 Words: Transformers for Image Recognition at Scale," Google Research, Brain Team, {adosovitskiy, neilhoulsby}@google.com, Published as a conference paper at ICLR 2021, 3 June 2021.
    
    \bibitem{b14} A. Kolesnikov, L. Beyer, X. Zhai, J. Puigcerver, J. Yung, S. Gelly, and N. Houlsby, "Big Transfer (BiT): General Visual Representation Learning," Google Research, Brain Team, Zürich, Switzerland, {akolesnikov, lbeyer, xzhai}@google.com, {jpuigcerver, jessicayung, sylvaingelly, neilhoulsby}@google.com, 5 May 2020.
    
    \bibitem{b15} M. Wortsman, G. Ilharco, S. Y. Gadre, R. Roelofs, R. Gontijo-Lopes, A. S. Morcos, H. Namkoong, A. Farhadi, Y. Carmon, S. Kornblith, and L. Schmidt, "Model Soups: Averaging Weights of Multiple Fine-tuned Models Improves Accuracy Without Increasing Inference Time," 1 July 2022.
    
    \bibitem{b16} M. Tan and Q. V. Le, "EfficientNet: Rethinking Model Scaling for Convolutional Neural Networks," 11 Sept 2020.

    \bibitem{b17} K. Maharana, S. Mondal, B. Nemade, "A review: Data pre-processing and data augmentation techniques," \textit{Global Transitions Proceedings}, vol. 3, no. 1, 2022, pp. 91-99, ISSN 2666-285X, https://doi.org/10.1016/j.gltp.2022.04.020.

    \bibitem{b18} L. Fei-Fei, R. Fergus, and P. Perona, "Learning generative visual models from few training examples: An incremental Bayesian approach tested on 101 object categories," \textit{Computer Vision and Pattern Recognition}, 2004. CVPR 2004. Proceedings of the 2004 IEEE Computer Society Conference on, vol. 1, I-988–I-995, 2004, https://doi.org/10.1109/CVPR.2004.1315239.

    \bibitem{b19} A. Vaswani, N. Shazeer, N. Parmar, J. Uszkoreit, L. Jones, A. N. Gomez, L. Kaiser, and I. Polosukhin, "Attention is All You Need," \textit{Advances in Neural Information Processing Systems 30 (NIPS 2017)}, Long Beach, CA, USA, December 4-9, 2017.

    \bibitem{b20} Alexander Kolesnikov, Lucas Beyer, Xiaohua Zhai, Joan Puigcerver, Jessica Yung, Sylvain Gelly, and Neil Houlsby, "Large Scale Learning of General Visual Representations for Transfer," \textit{CoRR}, vol. abs/1912.11370, 2019. [Online]. Available: \url{http://arxiv.org/abs/1912.11370}

    \bibitem{b21} Mitchell Wortsman, Gabriel Ilharco, Samir Yitzhak Gadre, Rebecca Roelofs, Raphael Gontijo-Lopes, Ari S. Morcos, Hongseok Namkoong, Ali Farhadi, Yair Carmon, Simon Kornblith, and Ludwig Schmidt, "Model Soups: Averaging Weights of Multiple Fine-tuned Models Improves Accuracy Without Increasing Inference Time," 2022. [Online]. Available: \url{https://arxiv.org/abs/2203.05482}

    \bibitem{b22} ImageNet, "ImageNet Large Scale Visual Recognition Challenge," \url{http://www.image-net.org/}.

    \bibitem{b23} E. D. Cubuk, B. Zoph, D. Mane, V. Vasudevan, and Q. V. Le, "AutoAugment: Learning Augmentation Policies from Data," CoRR, vol. abs/1805.09501, 2018, \url{http://arxiv.org/abs/1805.09501}.

    \bibitem{b24} Alex Krizhevsky, "Learning Multiple Layers of Features from Tiny Images," Tech. Rep., 2009, \url{https://www.cs.toronto.edu/~kriz/learning-features-2009-TR.pdf}.

    \bibitem{b25} Xingping Dai, Xiaoyu Zhao, Feng Cen, Fanglai Zhu, "Data Augmentation Using Mixup and Random Erasing," in \textit{2022 IEEE International Conference on Networking, Sensing and Control (ICNSC)}, 2022, pp. 1-6, doi: 10.1109/ICNSC55942.2022.10004083.

    \bibitem{b26} Pengguang Chen, Shu Liu, Hengshuang Zhao, Jiaya Jia, "GridMask Data Augmentation," 2020, arXiv:2001.04086 [cs.CV].

    \bibitem{b27} Terrance DeVries and Graham W. Taylor, "Improved Regularization of Convolutional Neural Networks with Cutout," 2017, arXiv:1708.04552 [cs.CV].

    \bibitem{b28} Xingping Dai, Xiaoyu Zhao, Feng Cen, and Fanglai Zhu, "Data Augmentation Using Mixup and Random Erasing," in \textit{2022 IEEE International Conference on Networking, Sensing and Control (ICNSC)}, 2022, pp. 1-6, doi: 10.1109/ICNSC55942.2022.10004083.

    \bibitem{b29} Y. Netzer, T. Wang, A. Coates, A. Bissacco, B. Wu, and A. Y. Ng, "Reading Digits in Natural Images with Unsupervised Feature Learning," in \textit{NIPS Workshop on Deep Learning and Unsupervised Feature Learning}, 2011.

    \bibitem{b30} Zhun Zhong, Liang Zheng, Guoliang Kang, Shaozi Li, and Yi Yang, "Random Erasing Data Augmentation," arXiv:1708.04896, 2017.


\end{thebibliography}
\end{document}